\pdfoutput=1

\documentclass[11pt]{article}

\usepackage{acl}

\usepackage{times}
\usepackage{latexsym}

\usepackage[T1]{fontenc}

\usepackage[utf8]{inputenc}

\usepackage{microtype}

\usepackage{inconsolata}

\usepackage{graphicx}
\usepackage{wrapfig}
\usepackage{algorithm}
\usepackage{algpseudocode}
\usepackage{array}
\usepackage{arydshln}
\usepackage{multirow}
\usepackage{makecell}
\usepackage{amsmath}
\usepackage[normalem]{ulem}
\usepackage{hhline}
\usepackage{amsfonts}

\usepackage{soul}

\usepackage[inkscapeexe=/Applications/Inkscape.app/Contents/MacOS/inkscape,inkscapearea=page]{svg}

\renewcommand{\hl}[1]{#1}

\usepackage{color}
\DeclareRobustCommand{\hlcyan}[1]{{\sethlcolor{cyan}\hl{#1}}}

\title{X-PEFT: eXtremely Parameter-Efficient Fine-Tuning \\ for Extreme Multi-Profile Scenarios}

\author{Namju Kwak \\
  Graduate School of Data Science \\
  Seoul National University \\
  \texttt{beatu@snu.ac.kr} \\\And
  Taesup Kim \\
  Graduate School of Data Science \\
  Seoul National University \\
  \texttt{taesup.kim@snu.ac.kr} \\}

\begin{document}
\maketitle
\begin{abstract}
Parameter-efficient fine-tuning (PEFT) techniques, such as adapter tuning, aim to fine-tune a pre-trained language model (PLM) using a minimal number of parameters for a specific task or profile. Although adapter tuning provides increased parameter efficiency compared to full-model fine-tuning, it introduces a small set of additional parameters attached to a PLM for each profile. This can become problematic in practical applications with multiple profiles, particularly when a significant increase in the number of profiles linearly boosts the total number of additional parameters. To mitigate this issue, we introduce X-PEFT, a novel PEFT method that leverages a multitude of given adapters by fine-tuning an extremely small set of compact tensors for a new profile, which serve as binary masks to adaptively select the given adapters. To efficiently validate our proposed method,
we implement it using a large number of \hlcyan{trained or untrained (random) adapters.}
We evaluate the performance of X-PEFT through \hlcyan{LaMP and} GLUE tasks and demonstrate that it either matches or surpasses the effectiveness of conventional adapter tuning, despite reducing the memory requirements per profile by a factor of 10,000 compared to it.
\end{abstract}

\section{Introduction}

Transfer learning, utilizing various pre-trained language models (PLMs) based on transformers \citep{devlin-etal-2019-bert, liu, lan, radford}, has demonstrated effectiveness across a wide range of NLP tasks.
Consequently, it has become common practice to fine-tune a PLM for a specific task rather than training a model from scratch.
However, as PLMs scale up, encompassing more than billions of parameters, a fine-tuning approach that updates all model parameters (i.e. full fine-tuning) becomes problematic.
For instance, fully fine-tuning a large PLM for a specific task not only requires more computational resources for training but also necessitates additional management of the fine-tuned large PLM.
Moreover, when this approach is repetitively applied for multiple tasks, the issues exacerbate.
To mitigate these issues, parameter-efficient fine-tuning (PEFT) approaches, such as adapter tuning \citep{houlsby, he} and prompt tuning \citep{li-liang-2021-prefix, lester-etal-2021-power}, have been proposed as alternatives to fine-tuning the entire model. These approaches are structured similarly, introducing a small number of additional parameters for each task and only fine-tuning those, and \citet{mao-etal-2022-unipelt} demonstrated that various PEFT approaches can be summarized within a unified view.

\newcommand\usesvg{0}

\begin{figure}
    \centering
    \ifnum\usesvg=0
        \includegraphics[width=1\linewidth]{fig_scalability.png}
    \else
        \def\svgwidth{\columnwidth}
        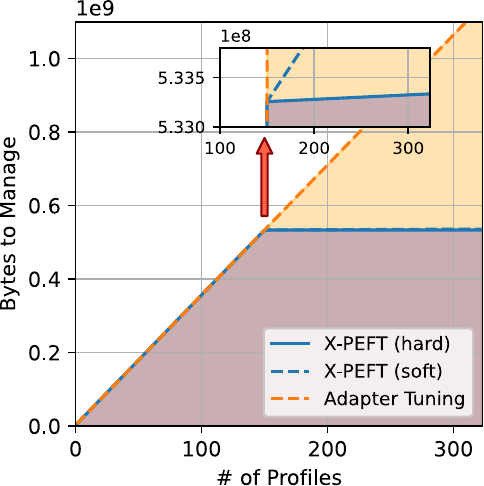
    \fi
    \vspace{-25pt}
    \caption{{\bf Demonstrating the remarkable parameter efficiency in terms of memory requirements of X-PEFT in extreme multi-profile scenarios.} 
    Additional details can be found in Section \ref{sec:lamp_exp}.
    }
    \label{fig:param-eff}
    \vspace{-15pt}
\end{figure}

Although PEFT approaches have achieved parameter efficiency with PLMs compared to full-model fine-tuning, practical scenarios, such as multi-profile applications that require management of a large number of profiles, demand even higher parameter efficiency. Consider, for instance, a scenario depicted in the LaMP \citep{salemi} dataset, where a news organization utilizes an article category classifier to assign articles to their respective categories, such as politics, economics, entertainment, and more. There may be numerous authors or profiles to manage, each with distinct criteria and preferences for categorizing articles, necessitating the management of a large number of sets of author-specific parameters (i.e. adapters). Another example is a personalized chatbot service, where the number of users may continually increase and user-specific parameters must be managed to effectively improve user experiences.

To further improve parameter efficiency in extreme multi-profile scenarios, we propose X-PEFT (eXtremely Parameter-Efficient Fine-Tuning), a novel PEFT method that leverages a multitude of given adapters by fine-tuning an extremely small set of compact tensors for a new profile, serving as binary masks to adaptively select the given adapters. As depicted in Figure~\ref{fig:param-eff}, \textit{our X-PEFT drastically reduces the memory requirements of additional parameters for each profile, by a factor of 10,000 compared to existing PEFT methods such as adapter tuning}. More specifically, after a certain number of profiles have been trained with adapter tuning and their adapter parameters have accumulated (e.g., 150 profiles in Figure~\ref{fig:param-eff}), each new incoming profile is designed to reuse and adaptively select them, rather than training a new adapter from scratch. Therefore, our proposed method only necessitates learning and storing binary mask tensors at the byte level, which are substantially more compact than adapters.

We conduct extensive experiments to validate our proposed method efficiently, using a large number of \hlcyan{trained and untrained (random)} adapters. Even with the use of \hlcyan{untrained} adapters, we demonstrate the proper applicability of X-PEFT. Interestingly, this can be viewed as an adapter-based version of the supermask concept~\citep{zhou}, aligning with the principles of the Lottery Ticket Hypothesis~\citep{frankle}. 

\begin{figure*}
    \centering
    \includegraphics[width=1.0\linewidth]{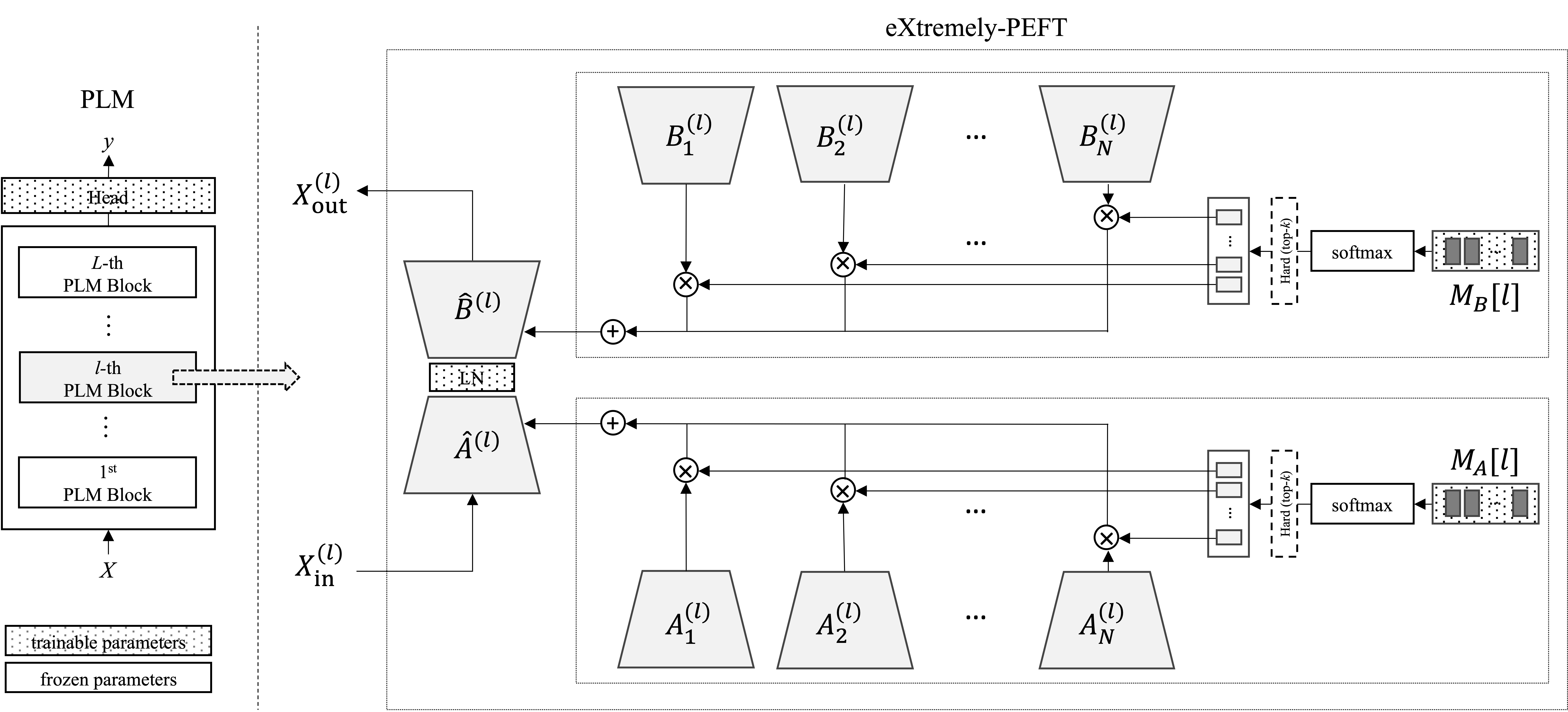}
    \caption{{\bf Illustration of our proposed method, X-PEFT}. Additional details can be found in Section~\ref{sec:method}.
    }
    \label{fig:archi}
\end{figure*}

Overall, our contributions can be summarized as follows:

\begin{itemize}
    \setlength\itemsep{0.0em}
    \item We introduce a novel PEFT method, X-PEFT, which achieves higher parameter efficiency by a factor of 10,000 compared to adapter tuning without sacrificing performance.
    \item We \hlcyan{implement X-PEFT using trained adapters and} apply it in a practical scenario using the LaMP dataset and demonstrate the efficiency of X-PEFT in terms of both performance and memory requirements.
    \item We \hlcyan{also} experimentally implement X-PEFT using a large number of \hlcyan{untrained (random)} adapters, based on the principles of the Lottery Ticket Hypothesis, and validate its effectiveness in extreme multi-profile scenarios.
\end{itemize}

\section{Multi-Profile Scenarios}

\hlcyan{
Multi-profile scenarios differ from single-profile counterparts in that modeling is required for multiple profiles. In the context of transfer learning, keeping the PLM frozen while fine-tuning for each profile is a reasonable choice. This necessitates the management of a set of parameters for the PLM and $P$ sets of profile-specific parameters, where $P$ denotes the number of profiles.

When $P$ is modest, the motivation to explore more parameter-efficient methods is limited. However, as $P$ increases, the motivation grows. At some point in the increasing $P$, the cumulative count of all profile-specific parameters may surpass that of PLM parameters. Consequently, existing X-PEFT methods are not parameter-efficient enough to accommodate highly multi-profile scenarios.

To elaborate, consider the case where $P$ is 1; in such instances, a singular set of profile-specific parameters suffices. Utilizing Pfeiffer adapters
}
\citep{pfeiffer-etal-2021-adapterfusion}
\hlcyan{
, this count amounts to 884.7K. Assuming our PLM is \texttt{bert-base-uncased}, with a parameter count of 110M, it exemplifies significant parameter efficiency. However, with $P$ set to 100, the total count of profile-specific parameters reaches 88.5M, approximately commensurate with the PLM parameter count of 110M. This lack of substantial parameter efficiency demands more efficient methods for fine-tuning.

Fewer profile-specific parameters mean less memory requirement to maintain profile-specific context for multi-profile applications. In multi-profile applications, profile-specific parameters should be adapted to the PLM as quickly as possible for quality of service. The reduction in the count of profile-specific parameters by X-PEFT is remarkable, diminishing it by a factor of 1/100. Furthermore, it significantly mitigates memory requirements per profile by a factor of 1/10,000, achieved through the judicious storage of profile-specific parameters.
}

\section{eXtremely-PEFT}\label{sec:method}
The goal of X-PEFT is to achieve parameter efficiency with PLMs, especially in extreme multi-profile scenarios where an increasing number of profiles continuously necessitates additional memory. For this reason, we propose to simply reuse a large number of given adapter, often in the hundreds, for each new profile rather than training additional new adapters from scratch for it. More specifically, we introduce some learnable mask tensors for each new profile, which are utilized to select and aggregate existing adapters into new ones during inference.

In terms of combining multiple adapters, X-PEFT is similar to AdapterFusion \citep{pfeiffer-etal-2021-adapterfusion}, but it combines existing adapters, which are fixed and shared across profiles, with lightweight mask tensors, thereby avoiding additional fusion layers. Moreover, as our primary focus is on extreme multi-profile scenarios, X-PEFT aims to effectively leverage knowledges  from a substantial number of existing adapters, as opposed to AdapterFusion, which utilizes a limited few.

In this work, we implement X-PEFT by defining adapters with the \hl{Pfeiffer configuration} \citep{pfeiffer-etal-2021-adapterfusion}. However, it is worth noting that X-PEFT can accommodate any other type of adapters \hl{such as LoRA} \citep{hu}. \hl{As a Pfeiffer adapter consists of two submodules, referred to as the down-projection and up-projection feed-forward networks,} denoted as $A$ and $B$ respectively, we define a model for each profile by adding a pair of $A^{(l)}$ and $B^{(l)}$ in each PLM block $l$. Based on this setting, we assume that $N$ adapters $\{(A_i^{(l)}, B_i^{(l)})\}_{i=1}^{N}$ for each PLM block $l$ are collected in advance by using
adapter tuning with $N$ profiles. After that, each new incoming profile is designed to adaptively select and reuse them instead of training new ones. 

\paragraph{X-PEFT with soft masks}
To implement this in a lightweight manner, we introduce two types of mask tensors for each new profile, namely $M_A$ and $M_B$. The former combines the submodule $A$'s of the adapters, while the latter combines the submodule $B$'s. Each mask tensor is a matrix $\mathbb{R}^{L \times N}$ in which each row corresponds to a mask for each PLM block, assigning different weights to $N$ adapters, subsequently utilized to construct new adapters $\hat{A}^{(l)}$ and $\hat{B}^{(l)}$ for each PLM block $l$ as follows:
\begin{equation*}
    \hat{A}^{(l)} = \sum_{i=1}^{N} M_A[l, i] A_i^{(l)}, \hat{B}^{(l)} = \sum_{i=1}^{N} M_B[l, i] B_i^{(l)},
\end{equation*}
where both adapters are applied to the input $X_{\text{in}}^{(l)}$ as $X_{\text{out}}^{(l)}=\hat{B}^{(l)}\hat{A}^{(l)}X_{\text{in}}^{(l)}$
\footnote{We insert layer normalization~(LN, \citet{ba}) after multiplying $\hat{A}^{(l)}$ that experimentally improved the overall performance.}
to compute the output $X_{\text{out}}^{(l)}$.
Here, the weight vectors $M_A[l]$ and $M_B[l]$ can be viewed as soft masks. In this approach, we treat each mask tensor as a regular trainable weight, similar to adapters but more compact, and apply the softmax activation before aggregating adapters to ensure that the weights sum to 1. This method is quite straightforward and does not require any special tricks to learn the mask tensors. Therefore, during fine-tuning for each new profile, we simultaneously and only optimize mask tensors and task header and freeze all other parameters related to PLM and $N$ adapters. In terms of memory requirements, it is more efficient than regular adapter tuning (see Table~\ref{tab:param_count}), but can be further improved by using hard masks instead of soft ones.

\paragraph{X-PEFT with hard masks}
By defining the mask tensors with hard masks, our method achieves significant parameter efficiency, particularly in terms of memory requirements, in extreme multi-profile scenarios. Hard masks require only binary masking information during inference. This means that for each new profile, we need to maintain only two bit arrays: one for $M_A$ and the other for $M_B$. 
We implement this by ensuring that the weight vectors $M_A[l]$ and $M_B[l]$ are $k$-hot vectors for all PLM layers, where  $k$ is the number of selected adapters within the given $N$ ones.  As $k$-hot vectors are non-differentiable, we employ the straight-through gradient estimation technique~\citep{bengio} for optimizing it through gumbel softmax~\citep{jang2017categorical, maddison2017the} with top-$k$ components (see Algorithm \ref{alg:softmax} in Appendix \ref{sec:app_alg}). Therefore, the mask tensors with hard masks are similarly optimized and utilized as the soft ones, except they are binarized into $k$-hot vectors after the training and stored in a more compact way.

\begin{table*}[t]
\centering
\begin{tabular}{c:cr:cr}
    \hline
    \multirow{2}{*}{\textbf{Mode}} & \multicolumn{2}{c:}{\textbf{Trainable Parameters}} & \multicolumn{2}{c}{\textbf{Memory Requirements}} \\
    \cline{2-5}
    & \multicolumn{1}{c}{\textbf{Formula}} & \multicolumn{1}{c:}{\textbf{Count}} & \multicolumn{1}{c}{\textbf{Formula}} & \multicolumn{1}{c}{\textbf{Byte}} \\
    \hline
    \multirow{3}{*}{\texttt{x\_peft} (hard)} & \multirow{6}{*}{$2(N+b) \times L$} & \multirow{2}{*}{${(N = 100)}$ 3.5K} & \multirow{3}{*}{$2 \lceil N / 8 \rceil \times L$} & ${(N=100)}$ 0.3K \\
    &  &  &  & ${(N=200)}$ 0.6K \\
    &  & \multirow{2}{*}{${(N = 200)}$ 5.9K} &  & ${(N=400)}$ 1.2K \\
    \cdashline{1-1}\cdashline{4-5}
    \multirow{3}{*}{\texttt{x\_peft} (soft)} &  &  & \multirow{3}{*}{$2N \times L \times 4$} & ${(N = 100)}$ 10K \\
    &  & \multirow{2}{*}{${(N = 400)}$ 10.7K} &  & ${(N = 200)}$ 20K \\
    &  &  &  & ${(N = 400)}$ 40K \\
    \hline
    \multicolumn{1}{c:}{\texttt{single\_adapter}} & \multicolumn{1}{c}{$2(d \times b) \times L$} & \multicolumn{1}{r:}{884.7K} & \multicolumn{1}{c}{$2(d\times b) \times L \times 4$} & \multicolumn{1}{r}{3.5M}   \\
    \hline
\end{tabular}
\caption{
{\bf Trainable parameters and memory requirements per profile.} In this comparison, we use a bottleneck dimension $b=64$, an adapter layer input dimension $d=768$, the number of PLM blocks $L=12$, and the number of given adapters $N=\{100, 200, 400\}$.  
}
\label{tab:param_count}
\end{table*}

\paragraph{Parameter efficiency}
The number of trainable parameters for each new profile with X-PEFT is calculated as $2(N + b) \times L$ \hlcyan{, where $b$ denotes the bottleneck dimension, and $L$ represents the number of PLM blocks}. This calculation includes two mask weight vectors and LN affine parameters across all PLM blocks, and it entirely depends on the given number of adapters, denoted by $N$. Conversely, a conventional adapter tuning method necessitates the complete training of submodules $A$ and $B$ for each profile, which is quantified as $2(d \times b) \times L$ \hlcyan{with $d$ representing the adapter layer input dimension}.  As demonstrated in Table~\ref{tab:param_count}, the number of trainable parameters with X-PEFT remains constant regardless of the mask type and can be substantially reduced, by a factor of around 100, even when the number of adapters $N$ is increased up to 400.

More interestingly, if we focus on the memory requirements to store these trainable parameters for each profile, X-PEFT with hard masks can further improve the parameter efficiency by a factor of around 10,000 compared to adapter tuning. This ultimately shows how X-PEFT can be used efficiently in extreme multi-profile scenarios.

\paragraph{Connection to Lottery Ticket Hypothesis} \label{sec:lth}
The Lottery Ticket Hypothesis~\citep{frankle} asserts that a randomly-initialized dense neural network contains a subnetwork, initialized in a manner that enables it to achieve test accuracy comparable to that of the original network after training, for at most the same number of iterations. Furthermore, \citet{zhou} introduced the concept of a supermask, represented as a weight-level mask, which can deliver better-than-chance test accuracy without any training. Similarly, yet distinctively, X-PEFT involves searching for a supermask within a set of given adapters. This search aims to discover adapter-level masks instead of weight-level ones. In other words, our masking granularity corresponds to an entire adapter, whereas \citet{zhou}'s supermask operates at the level of individual weights. 

Based on this interpretation, we validate our method with a large number of \hlcyan{untrained random} adapters rather than \hlcyan{trained} ones. This allows us to efficiently simulate X-PEFT in extreme multi-profile scenarios by effortlessly increasing the number of adapters $N$ with random adapters\hlcyan{. This use of random adapters posits them as proper proxies for hypothethical pre-existing $N$ profiles, and our objective is to discern optimal mask tensors within this simulated ensemble of profiles.} Even with \hlcyan{these} random adapters, our experimental results indicate that X-PEFT can function effectively without any severe performance degradation (see details in Section~\ref{sec:results}). Moreover, it is noteworthy that, even when employing entirely different sets of random adapters, performance is consistently guaranteed, as shown in Figure \ref{fig:random_seed} in Appendix \ref{sec:app_fig}.

\begin{table*}[ht!]
    \centering
    \begin{tabular}{ccccccccccccc}
        \hline
        \textbf{Mode} & \textbf{Adapters} & \textbf{\texttt{cola}} & \textbf{\texttt{sst2}} & \textbf{\texttt{mrpc}} & \textbf{\texttt{qqp}} & \textbf{\texttt{stsb}} & \textbf{\texttt{mnli}} & \textbf{\texttt{qnli}}  & \textbf{\texttt{rte}} & \textbf{\texttt{wnli}}\\
        & & \scriptsize{(MCC)} & \scriptsize{(Acc)} & \scriptsize{(Comb)} & \scriptsize{(Comb)} & \scriptsize{(Comb)} & \scriptsize{(Comb)} & \scriptsize{(Acc)} & \scriptsize{(Acc)} & \scriptsize{(Acc)} \\
        \hline
        \texttt{x\_peft} & 100 \small{(soft)} & 0.40 & 0.90 & 0.78 & 0.79 & 0.79 & 0.68 & 0.82 & 0.58 & 0.34\\
        & 100 \small{(hard)} & 0.39 & 0.87 & 0.76 & 0.76 & 0.74 & 0.63 & 0.76 & \underline{\textbf{0.61}} & 0.32\\
        & 200 \small{(soft)} & 0.44 & \underline{0.91} & 0.78 & 0.80 & 0.80 & 0.69 & 0.83 & 0.60 & \underline{0.37}\\
        & 200 \small{(hard)} & 0.44 & 0.89 & 0.81 & 0.77 & 0.76 & 0.65 & 0.79 & 0.58 & 0.34\\
        & 400 \small{(soft)} & \underline{\textbf{0.47}} & 0.90 & 0.78 & \underline{0.81} & 0.81 & \underline{0.72} & \underline{0.83} & 0.58 & 0.30\\
        & 400 \small{(hard)} & 0.46 & 0.89 & \underline{\textbf{0.82}} & 0.78 & \underline{\textbf{0.81}} & 0.67 & 0.81 & 0.55 & 0.27\\
        \hdashline
        \texttt{head\_only} & - & 0.31 & 0.85 & 0.76 & 0.72 & 0.46 & 0.53 & 0.68 & 0.59 & 0.38\\
        \hdashline
        \texttt{single\_adapter} & - & 0.43 & \textbf{0.91} & 0.76 & \textbf{0.85} & 0.80 & \textbf{0.80} & \textbf{0.88} & 0.60 & \textbf{0.42}\\
        \hline
    \end{tabular}
    \caption{{\bf Evaluation of the GLUE tasks.} In the case of hard masking, we employ $k = 50$. The scores in the table are reported based on the official metrics provided by the GLUE dataset. When multiple official metrics exist for a task (indicated as `Comb'), we present the combined score (i.e., mean). `Acc' and `MCC' denote accuracy and Matthew's Correlation, respectively. The full individual metric data can be found in the Appendix \ref{sec:app_eval}. Underlined values represent the best among \texttt{x\_peft} cases, and bold-faced values represent the best among all three modes.    
    }
    \label{tab:glue_eval}
\end{table*}

\section{Experiments}
We evaluate the effectiveness of our proposed method, X-PEFT, by conducting experiments across a wide variety of settings and comparing it against baselines:
\begin{itemize}
    \setlength\itemsep{0.0em}
    \item \texttt{x\_peft} (\texttt{xp}): Our proposed method, X-PEFT, with mask tensors.    
    \item \texttt{single\_adapter} (\texttt{sa}): Standard adapter tuning with a single adapter.
    \item \texttt{head\_only} (\texttt{ho}): Fine-tuning only the downstream without any adapter.
\end{itemize}

\paragraph{Experimental settings}
In all experiments, we employ \texttt{bert-base-uncased} \citep{devlin-etal-2019-bert} as the PLM and \hl{Pfeiffer adapter} \citep{pfeiffer-etal-2021-adapterfusion} with a reduction factor $r=16$ (bottleneck dimension $b=48$) for adapters.
We set a random seed of 42 for all experiments and conduct a separate experiment to verify reproducibility by varying the random seed (see Figure \ref{fig:random_seed} in Appendix \ref{sec:app_fig}).
The AdamW optimizer is used with a learning rate of $1.0 \times 10^{-05}$, which underwent linear decay, and the training duration for all experiments is set to 10 epochs.
We use 4 GPUs (GeForce RTX 3090) and exploit data parallelism for all experiments. Moreover, we apply gradient checkpointing \citep{chen} to improve computational efficiency of \texttt{x\_peft}.
All experimental cases are given an equal number of training samples to maintain fairness. We use a consistent batch size across all experiments to ensure that parameters received an equal number of updates. 
The majority of experiments employ a batch size of 64 and a token sequence length of 128. \hl{However, in the case of $N = 800$, a batch size of 32 is applied.}
For implementation, we used the Hugging Face's \texttt{transformers} \citep{wolf-etal-2020-transformers} and AdapterHub's \texttt{adapter-transformers} \citep{pfeiffer-etal-2020-adapterhub} packages.

\paragraph{Datasets}

We conduct experiments using the LaMP \citep{salemi} benchmark. However, since LaMP is originally designed for prompt tuning, modifications were necessary for our purposes. Further details regarding these modifications can be found in the Appendix \ref{sec:app_lamp_mod}. In particular, we utilize the `Personalized News Categorization' dataset from LaMP.

To \hlcyan{evaluate X-PEFT for the general single-profile settings and simulated multi-profile scenarios}, we incorporate 9 tasks from the GLUE benchmark \citep{wang} and 4 tasks from the more challenging SuperGLUE benchmark \citep{wang-superglue}. Our choice of SuperGLUE tasks (\texttt{cb}, \texttt{boolq}, \texttt{axb}, and \texttt{axg}) aligns with our use of the \texttt{bert-base-uncased} model, which supports single-sentence and sentence-pair input formats. We conduct evaluation for GLUE and SuperGLUE on the development (\texttt{dev}) sub-dataset using the evaluation metrics officially suggested by the benchmark. For the \texttt{axg} task in SuperGLUE, we also utilize the Winogender \citep{rudinger-etal-2018-gender} test dataset recast by \citet{poliak-etal-2018-collecting}.

\subsection{Experimental Results} \label{sec:results}

\hlcyan{The LaMP experiments exemplify the efficacy of X-PEFT with trained adapters, showcasing significantly enhanced parameter efficiency tailored for multi-profile scenarios.}

\hlcyan{In parallel, the GLUE and SuperGLUE experiments underscore the inherent capabilities of the X-PEFT architecture. These experiments involve assessments of X-PEFT with untrained random adapters, providing insights into its performance in general single-profile settings.}

\hlcyan{Additionally, the GLUE and SuperGLUE experiments can be interpreted as simulated scenarios for multi-profile settings. This perspective arises when considering X-PEFT with random adapters as a mechanism for identifying adapter-level supermasks.}

\paragraph{LaMP with \hlcyan{trained} adapters} \label{sec:lamp_exp}
Our modified LaMP dataset follows the schema (\texttt{news\_text}, \texttt{news\_category}, \texttt{author\_profile}), structuring each data point for text classification while incorporating the author's identity. It contains 17,005 news texts authored by 323 individuals / profiles. On average, each author contributed 52.65 news texts, with a standard deviation of 87.28 (ranging from a minimum of 6 to a maximum of 640).

\begin{figure}[ht]
    \centering
    \ifnum\usesvg=0
        \includegraphics[width=1\linewidth]{fig_lamp_evaluation.png}
    \else
        \def\svgwidth{\columnwidth}
        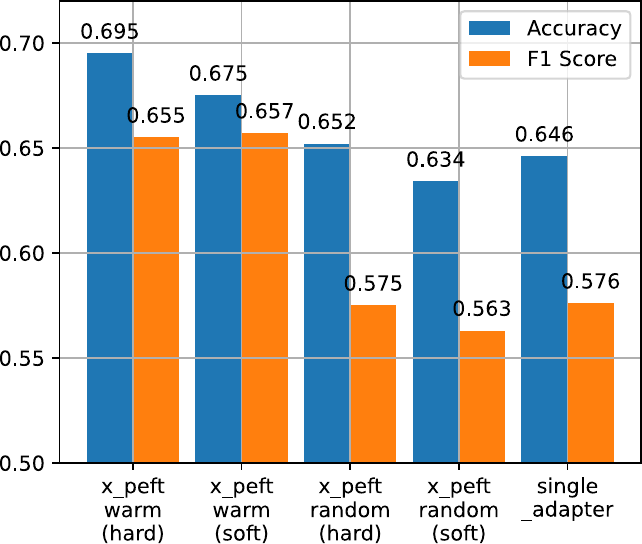
    \fi
    \caption{{\bf Evaluation of the Modified LaMP `Personalized News Categorization' Dataset.} Averaged evaluation accuracy and F1 score over 323 authors are presented (on 30\% holdout sets).}
    \label{fig:enter-label}
\end{figure}

In our experiment for \texttt{x\_peft}, \hl{we generate 150 random adapters (denoted by \texttt{x\_peft random}) which are frozen and shared across 323 authors. For each profile, utilizing aforementioned 150 adapters, mask tensors are optimized individually (per-profile). As a result, we obtain a collection of mask tensors for 323 authors.} These mask tensors serve as a highly efficient means of personalizing the model for multi-profile scenarios. 
The averaged evaluation accuracy and F1 scores for all 323 authors can be found in Figure~\ref{fig:enter-label}, 
where \hl{\texttt{x\_peft random}} with hard masks shows improvement compared to \texttt{single\_adapter}.

Moreover, we conduct additional experiments for \texttt{x\_peft}. \hl{We assume that the first 150 authors can 
train 150 adapters instead of keeping them frozen as random, as part of the warm-start procedure.} 
Subsequently, we conduct individual (per-profile) training for 173 authors by optimizing mask tensors using those 150 learned adapters. In this approach, \texttt{x\_peft} only has to train mask tensors and reuse the shared head without any fine-tuning. This setting (\hl{\texttt{x\_peft warm}}) can not only further improve the parameter efficiency than the previous setting (\hl{\texttt{x\_peft random}}), but also significantly improve the averaged evaluation accuracy and F1 scores as depicted in Figure~\ref{fig:enter-label}.

\begin{figure}[ht]
    \centering
    \includegraphics[width=1\linewidth]{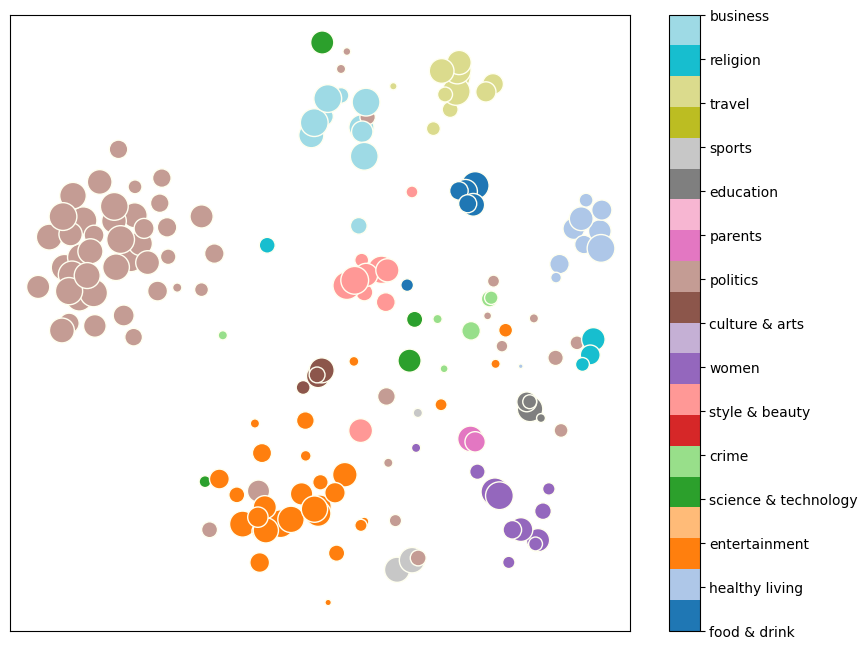}
    \caption{{\bf Visualization of mask tensors with t-SNE}. Each point represents an author/profile, and the color and size of it represent the majority category assigned by each author and the majority ratio in an article. This shows how the mask tensors effectively capture the categorization diversity among authors.}
    \label{fig:lamp_tsne}
\end{figure}

\begin{figure}
    \centering
    \ifnum\usesvg=0
        \includegraphics[width=1\linewidth]{fig_lamp_heatmap.png}
    \else
        \def\svgwidth{\columnwidth}
        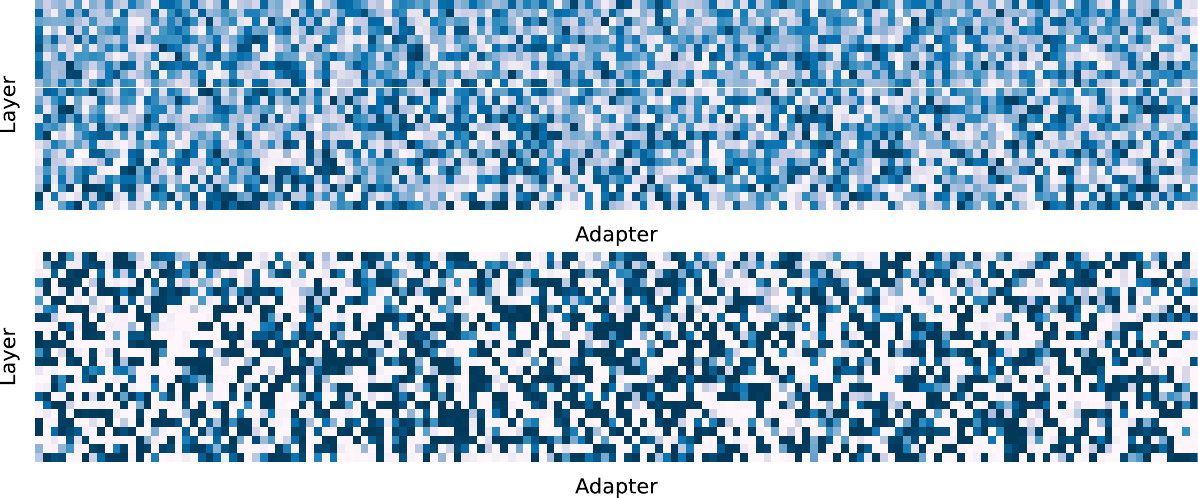
    \fi
    \caption{{\bf Heatmaps for mask tensors of most distant authors.} These distinct heatmaps capture the unique characteristics of news categorization.
    }
    \label{fig:lamp_heatmap}
\end{figure}

The memory requirements for this setting are precisely shown in Figure~\ref{fig:param-eff}. Essentially, these mask tensors encapsulate a unique signature of each author, specifically revealing how they categorize news texts in this context. We visualize the 173 sets of mask tensors using t-SNE \citep{maaten} in Figure \ref{fig:lamp_tsne}, along with heatmaps illustrating the mask tensors of the two most distant (Euclidean) profiles as shown in Figure \ref{fig:lamp_heatmap}.

\paragraph{GLUE with \hlcyan{untrained} adapters}
We conduct experiments on 9 tasks using their respective datasets. For the \texttt{x\_peft} case, we utilize 100, 200, and 400 random adapters, to efficiently validate its effectiveness, each for both soft and hard masking. As baselines, we experiment the \texttt{single\_adapter} and \texttt{head\_only} cases. The overall results are presented in Table \ref{tab:glue_eval}.

Our expectation was that the best evaluation score achieved by any \texttt{x\_peft} experiment for each task would fall between the evaluation scores of \texttt{head\_only} and \texttt{single\_adapter}. \texttt{head\_only} represents the lower bound, as \texttt{x\_peft} involves training a downstream head in addition to mask tensors. On the other hand, \texttt{single\_adapter} represents the upper bound, as our objective was to demonstrate that \texttt{x\_peft} could achieve comparable or superior performance to \texttt{single\_adapter} with significantly fewer trainable parameters.

As expected, for all tasks except \texttt{wnli}, \texttt{x\_peft}'s best evaluation scores exceed those of \texttt{head\_only}. Unexpectedly, in about half of the tasks, \texttt{x\_peft} even outperforms \texttt{single\_adapter} in terms of evaluation score. For tasks where \texttt{x\_peft} falls between \texttt{head\_only} and \texttt{single\_adapter}, it is much closer to \texttt{single\_adapter} in performance with a negligible gap. It is noteworthy given the significantly lower number of trainable parameters in \texttt{x\_peft} in comparison to \texttt{single\_adapter}. 

\paragraph{SuperGLUE with \hlcyan{untrained} adapters}
We conduct experiments on tasks \texttt{cb} and \texttt{boolq} using their respective datasets. Additionally, we perform experiments on diagnostic tasks \texttt{axb} and \texttt{axg} using GLUE's \texttt{rte} dataset for training. For the \texttt{x\_peft} case, we apply the same setting of using random adapters as the evaluation of GLUE tasks. The overall results are presented in Table \ref{tab:superglue_eval}.

Similarly to the evaluation results of the GLUE experiments, in all cases, \texttt{x\_peft}'s highest evaluation scores match or even surpass those of \texttt{single\_adapter}. Unexpectedly, for \texttt{cb}, \texttt{head\_only} performs the best. This performance from \texttt{x\_peft} is noteworthy considering the significantly lower number of trainable parameters.

\begin{table}[t]
    \centering
    \begin{tabular}{cccccccc}
        \hline
        \textbf{M.} & \textbf{Adt.} & \textbf{\texttt{cb}} & \scalebox{0.8}{\textbf{\texttt{boolq}}} & \textbf{\texttt{axb}} & \textbf{\texttt{axg}} & \textbf{\texttt{axg}}\\
        & & \scriptsize{(Acc)} & \scriptsize{(Acc)} & \scriptsize{(MCC)} & \scriptsize{(Acc)} & \scriptsize{(GPS)}\\
        \hline
        \texttt{xp} & 100 \small{(s)} & .64 & .67 & .11 & \textbf{\underline{.53}} & 92.7\\
        & 100 \small{(h)} & .68 & .66 & .09 & .48 & 86.6\\
        & 200 \small{(s)} & .68 & .66 & .07 & .52 & \textbf{\underline{96.1}}\\
        & 200 \small{(h)} & .68 & .66 & .02 & .50 & 88.4\\
        & 400 \small{(s)} & .68 & .66 & .09 & .51 & 93.5\\
        & 400 \small{(h)} & \underline{.70} & \textbf{\underline{.68}} & \textbf{\underline{.12}} & .50 & 94.8\\
        \hdashline
        \texttt{ho} & - & \textbf{.71} & .64 & .09 & .50 & 82.3\\
        \hdashline
        \texttt{sa} & - & .68 & .65 & .10 & .51 & 93.5\\
        \hline
    \end{tabular}
    \caption{{\bf Evaluation of the SuperGLUE tasks.} For hard masking, we employ $k = 50$. `GPS' denotes Gender Parity Score, and all other symbols and text decorations can be understood in the same context as in Table \ref{tab:glue_eval}.}
    \label{tab:superglue_eval}
\end{table}

\begin{figure*}[ht]
    \centering
    \ifnum\usesvg=0
        \includegraphics[width=\linewidth]{fig_analysis.png}
    \else
        \def\svgwidth{\linewidth}
        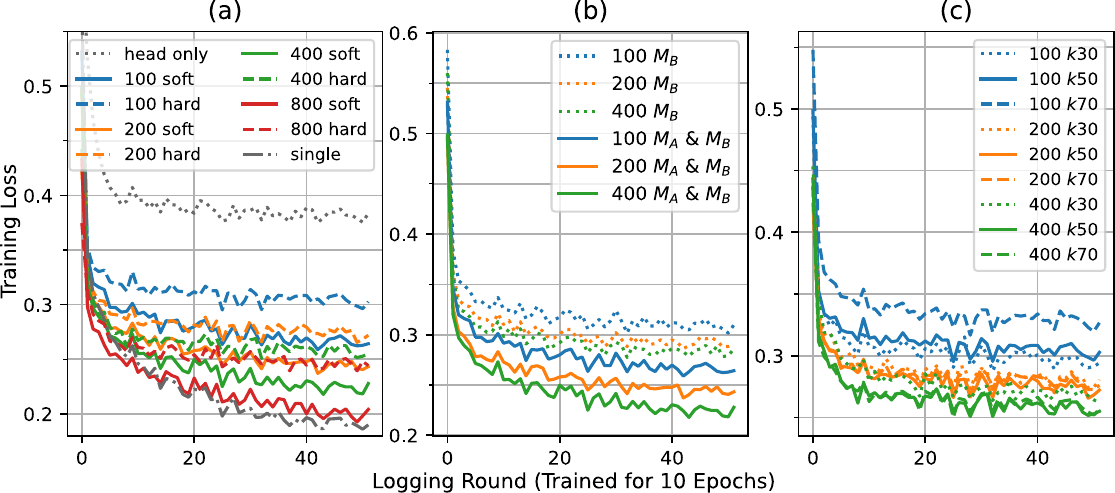
    \fi
    \caption{{\bf Training curves for \texttt{sst2} with various settings.} (a) Varying the number of adapters and comparing soft / hard masks: more adapters lead to improved loss, and soft masks generally show lower loss than hard ones. (b) Effectiveness of separate mask tensors: the impact of having $M_A$ and $M_B$ is evident. (c) Varying $k$ for hard masks: $k = 50$ consistently shows best performance irrespective of the specific value of $N$.
    }
    \label{fig:plots}
\end{figure*}

\subsection{Ablation Studies and Analysis}

\paragraph{The number of given adapters ($N$)}
When analyzing training curves, X-PEFT with a higher number of adapters outperforms its counterparts. As shown in Figure \ref{fig:plots} (a), the training curves for the \texttt{sst2} task consistently position lower when more adapters are used. This trend is consistent across various tasks in the GLUE benchmark.

In terms of evaluation scores, utilizing a higher number of adapters generally corresponds to better evaluation performance, even though they are random adapters, as demonstrated in Table \ref{tab:glue_eval}. However, there are some exceptions, notably in tasks such as \texttt{rte}, \texttt{sst2}, and \texttt{wnli}, where an abundance of adapters can potentially lead to overfitting.

\paragraph{Soft masks vs. hard masks}
We introduce X-PEFT with mask tensors, which can be implemented by either soft or hard masks. Each type has its own advantages and disadvantages. To validate these observations, we compare the two settings across our experiments. In most experiments, X-PEFT with hard masks demonstrated superior generalization performance compared to the soft ones (refer to Table~\ref{tab:glue_eval} and Figure~\ref{fig:enter-label}).  However, as depicted in Figure~\ref{fig:plots} (a), soft masks consistently display a lower training loss than their hard ones. From these results, we infer that soft masks are more prone to overfitting, whereas hard masks enhance generalization capabilities.

\paragraph{Separate mask tensors for submodules}
How can a small number of additional trainable parameters (e.g., mask tensor) in X-PEFT achieve performance that matches adapter tuning? The key factor is the use of two mask tensors instead of just one. When we use only one mask tensor for aggregating adapters (i.e., including only $M_B$ and discarding $M_A$ from the bottleneck), the expressive capacity for a new adapter is limited to $N$. However, when we use $M_A$ and $M_B$ together in sequence (i.e., separate mask tensors for submodules $A$ and $B$) can express $N^2$ cases. We conducted experiments to investigate this aspect on the \texttt{sst2} task as shown in Figure \ref{fig:plots} (b). Through these experiments, the results confirm that the combination of these two tensors can improve the performance of X-PEFT.

\paragraph{Top-$k$ selection for hard masks}
For X-PEFT with hard masks, $k = 50$ typically yields favorable training performances. For $N = 200$ and $N = 400$, increasing $k$ improves training performance until reaching $k = 50$, after which it begins to decline. For $N = 100$, a training performance peak is observed at $k = 30$, deviating from the pattern observed with larger $N$ values. Refer to Figure \ref{fig:plots} (c) for further insights. As $k$ diverges further from the value of $k = 50$, the loss curves progressively deviate from the curve of $k = 50$. In general, regardless of the value of $N$, $k = 50$ proves to be a highly reasonable choice.

\section{Related Works}
AdapterFusion \citep{pfeiffer-etal-2021-adapterfusion} is similar to X-PEFT in using multiple adapters, but it relies on attention tensors (referred to as fusion layers) for combining adapters, making it computationally heavy. This characteristic limits the scalability of AdapterFusion in multi-profile scenarios.

Parallel adapters \citep{ruckle-etal-2021-adapterdrop} and its scaled variant \citep{he} are essentially implementations of AdapterFusion, designed to enable parallel computation throughout adapters. Consequently, they inherit the limitations associated with AdapterFusion.
AdapterDrop \citep{ruckle-etal-2021-adapterdrop} is built upon the AdapterFusion architecture but focuses on pruning less significant adapters based on their activation strength. However, the number of adapters to be pruned varies from task to task, and it still employs attention tensors, making it computationally intensive.

AdapterSoup \citep{chronopoulou-etal-2023-adaptersoup} trains a set of adapters and selects a subset of them for inference. As this subset selection occurs at test time, the selection changes with different inputs, requiring the retention of all trained adapters at test time, which is not scalable.

AdaMix \citep{wang-etal-2022-adamix} employs the Mixture-of-Experts concept in adapter tuning. It trains a route policy among layers of adaptation modules, which comprise a mixture of adapters. It combines the weights of adaptation modules selected by an input batch for test time efficiency but requires the retention of all the trained adapters for inference, making it less scalable.

\citet{wu-etal-2022-pruning} employ the original approach proposed by the Lottery Ticket Hypothesis \citep{frankle}, applied at the adapter level within the AdapterFusion configuration. They iteratively prune a portion of the adapters until winning tickets are discovered. In contrast, X-PEFT can be viewed as the process of identifying supermasks \citep{zhou} among parallelly arranged adapter submodules.

Including the aforementioned works, as far as we know, X-PEFT is the first to involve hundreds of adapters (up to 800), except for \citet{wu-etal-2022-pruning}, which uses a maximum of 192 adapters (while others use fewer than 100 adapters). Additionally, as far as we know, X-PEFT is the first to apply the supermask \citep{zhou} concept in PEFT, particularly for multi-profile scenarios.

\hlcyan{
The aforementioned works also primarily concentrate on achieving heightened performance through the strategic composition and architecture of pre-existing building blocks, largely without delving into considerations of extreme parameter efficiency. In contrast, X-PEFT is distinctly oriented towards the reduction of parameters to an extraordinary degree, not sacrificing performance. In our pursuit of methodological effectiveness, we advocate for the utilization of the simplest building blocks. This rationale underscores our selection of the Pfeiffer adapter
}
\citep{pfeiffer-etal-2021-adapterfusion}
\hlcyan{
as both a foundational component and a benchmark for our study.
}

\section{Conclusion}
In this paper, we introduce eXtremely-PEFT, X-PEFT, a groundbreaking approach to Parameter-Efficient Fine-Tuning for pre-trained language models (PLMs). Our work achieves an unprecedented 1/100 reduction in parameters compared to adapter tuning while maintaining task performance. We also optimize the memory requirements, minimizing them to the byte level by a factor of 10,000, which is crucial for extreme multi-profile scenarios.

Furthermore, we delve deeper into PEFT, significantly reducing trainable parameters, thus reducing resource and computational costs. By incorporating the principles of the Lottery Ticket Hypothesis into adapter-level PEFT, X-PEFT opens new possibilities for resource-efficient natural language processing with PLMs.

Our work not only advances PEFT but also sets the stage for future research in NLP, inspiring novel applications and resource-efficient natural language processing breakthroughs. In conclusion, X-PEFT is a transformative development in PEFT, offering remarkable parameter efficiency without performance compromise. 

\newpage
\section*{Limitations}
Due to the extensive number of adapters involved in X-PEFT, training can be time-consuming. For hard masking, it is possible to reduce training time by disabling gradients for out-of-top-$k$ adapter submodules. We can also explore concepts from Parallel adapters \citep{ruckle-etal-2021-adapterdrop} about parallel computation for AdapterFusion \citep{pfeiffer-etal-2021-adapterfusion}.

There are almost no datasets available for multi-profile benchmarking. LaMP \citep{salemi} is currently the only dataset that exists for such purposes, but it is primarily designed for prompt tuning. While we did conduct multi-profile experiments on LaMP, these experiments necessitated some modifications. Unfortunately, the scarcity of multi-profile benchmark datasets limited our ability to carry out more comprehensive multi-profile experiments.

Regarding language, our research is constrained to English texts. The PLM utilized in our study has been specifically trained in English, and the datasets we employed are also in English. Future work will need to explore an extended approach to enhance parameter efficiency and multi-profile scalability, especially for low-resource languages.

\section*{Ethics Statement}
Gender bias in NLP models is a serious problem that can have far-reaching consequences, potentially undermining social integration and peace. Researchers in the NLP field have a responsibility to consider and address potential gender bias in all their efforts. The SuperGLUE benchmark includes a diagnostic dataset focusing on gender bias, namely \texttt{axg}. This is why we have included SuperGLUE in our experimental dataset.

X-PEFT offers the advantage of enabling multi-profile service providers to operate with minimal memory or storage requirements, ultimately reducing the strain on data centers and contributing to a reduction in carbon dioxide emissions. However, it's worth noting that the extended training times involving multiple GPUs can be environmentally problematic. Therefore, our ongoing research efforts are dedicated to achieving more efficient training methods and conserving computational power from an ecological viewpoint.

\bibliography{xpeft}

\begin{thebibliography}{30}
\expandafter\ifx\csname natexlab\endcsname\relax\def\natexlab#1{#1}\fi

\bibitem[{Ba et~al.(2016)Ba, Kiros, and Hinton}]{ba}
Jimmy~Lei Ba, Jamie~Ryan Kiros, and Geoffrey~E. Hinton. 2016.
\newblock \href {https://arxiv.org/abs/1607.06450} {Layer normalization}.
\newblock In \emph{arXiv:1607.06450 [stat.ML]}.

\bibitem[{Bengio et~al.(2013)Bengio, Léonard, and Courville}]{bengio}
Yoshua Bengio, Nicholas Léonard, and Aaron Courville. 2013.
\newblock \href {https://arxiv.org/abs/1308.3432} {Estimating or propagating gradients through stochastic neurons for conditional computation}.
\newblock In \emph{arXiv:1308.3432 [cs.LG]}.

\bibitem[{Chen et~al.(2016)Chen, Xu, Zhang, and Guestrin}]{chen}
Tianqi Chen, Bing Xu, Chiyuan Zhang, and Carlos Guestrin. 2016.
\newblock \href {https://arxiv.org/abs/1604.06174} {Training deep nets with sublinear memory cost}.
\newblock In \emph{arXiv:1604.06174 [cs.LG]}.

\bibitem[{Chronopoulou et~al.(2023)Chronopoulou, Peters, Fraser, and Dodge}]{chronopoulou-etal-2023-adaptersoup}
Alexandra Chronopoulou, Matthew Peters, Alexander Fraser, and Jesse Dodge. 2023.
\newblock \href {https://doi.org/10.18653/v1/2023.findings-eacl.153} {{A}dapter{S}oup: Weight averaging to improve generalization of pretrained language models}.
\newblock In \emph{Findings of the Association for Computational Linguistics: EACL 2023}, pages 2054--2063, Dubrovnik, Croatia. Association for Computational Linguistics.

\bibitem[{Devlin et~al.(2019)Devlin, Chang, Lee, and Toutanova}]{devlin-etal-2019-bert}
Jacob Devlin, Ming-Wei Chang, Kenton Lee, and Kristina Toutanova. 2019.
\newblock \href {https://doi.org/10.18653/v1/N19-1423} {{BERT}: Pre-training of deep bidirectional transformers for language understanding}.
\newblock In \emph{Proceedings of the 2019 Conference of the North {A}merican Chapter of the Association for Computational Linguistics: Human Language Technologies, Volume 1 (Long and Short Papers)}, pages 4171--4186, Minneapolis, Minnesota. Association for Computational Linguistics.

\bibitem[{Frankle and Carbin(2019)}]{frankle}
Jonathan Frankle and Michael Carbin. 2019.
\newblock \href {https://arxiv.org/abs/1803.03635} {The lottery ticket hypothesis: Finding sparse, trainable neural networks}.
\newblock In \emph{International Conference on Learning Representations}.

\bibitem[{He et~al.(2022)He, Zhou, Ma, Berg-Kirkpatrick, and Neubig}]{he}
Junxian He, Chunting Zhou, Xuezhe Ma, Taylor Berg-Kirkpatrick, and Graham Neubig. 2022.
\newblock \href {https://openreview.net/forum?id=0RDcd5Axok} {Towards a unified view of parameter-efficient transfer learning}.
\newblock In \emph{International Conference on Learning Representations}.

\bibitem[{Houlsby et~al.(2019)Houlsby, Giurgiu, Jastrzebski, Morrone, Laroussilhe, Gesmundo, Attariyan, and Gelly}]{houlsby}
Neil Houlsby, Andrei Giurgiu, Stanislaw Jastrzebski, Bruna Morrone, Quentin~De Laroussilhe, Andrea Gesmundo, Mona Attariyan, and Sylvain Gelly. 2019.
\newblock \href {http://proceedings.mlr.press/v97/houlsby19a.html} {Parameter-efficient transfer learning for nlp}.
\newblock In \emph{Proceedings of the 36th International Conference on Machine Learning, PMLR 97:2790-2799, 2019.}

\bibitem[{Hu et~al.(2022)Hu, Shen, Wallis, Allen-Zhu, Li, Wang, Wang, and Chen}]{hu}
Edward~J. Hu, Yelong Shen, Phillip Wallis, Zeyuan Allen-Zhu, Yuanzhi Li, Shean Wang, Lu~Wang, and Weizhu Chen. 2022.
\newblock \href {https://arxiv.org/abs/2106.09685} {Lora: Low-rank adaptation of large language models}.
\newblock In \emph{International Conference on Learning Representations}.

\bibitem[{Jang et~al.(2017)Jang, Gu, and Poole}]{jang2017categorical}
Eric Jang, Shixiang Gu, and Ben Poole. 2017.
\newblock \href {https://openreview.net/forum?id=rkE3y85ee} {Categorical reparameterization with gumbel-softmax}.
\newblock In \emph{International Conference on Learning Representations}.

\bibitem[{Lan et~al.(2020)Lan, Chen, Goodman, Gimpel, Sharma, and Soricut}]{lan}
Zhenzhong Lan, Mingda Chen, Sebastian Goodman, Kevin Gimpel, Piyush Sharma, and Radu Soricut. 2020.
\newblock \href {https://arxiv.org/abs/1909.11942} {Albert: A lite bert for self-supervised learning of language representations}.
\newblock In \emph{International Conference on Learning Representations}.

\bibitem[{Lester et~al.(2021)Lester, Al-Rfou, and Constant}]{lester-etal-2021-power}
Brian Lester, Rami Al-Rfou, and Noah Constant. 2021.
\newblock \href {https://doi.org/10.18653/v1/2021.emnlp-main.243} {The power of scale for parameter-efficient prompt tuning}.
\newblock In \emph{Proceedings of the 2021 Conference on Empirical Methods in Natural Language Processing}, pages 3045--3059, Online and Punta Cana, Dominican Republic. Association for Computational Linguistics.

\bibitem[{Li and Liang(2021)}]{li-liang-2021-prefix}
Xiang~Lisa Li and Percy Liang. 2021.
\newblock \href {https://doi.org/10.18653/v1/2021.acl-long.353} {Prefix-tuning: Optimizing continuous prompts for generation}.
\newblock In \emph{Proceedings of the 59th Annual Meeting of the Association for Computational Linguistics and the 11th International Joint Conference on Natural Language Processing (Volume 1: Long Papers)}, pages 4582--4597, Online. Association for Computational Linguistics.

\bibitem[{Liu et~al.(2020)Liu, Ott, Goyal, Du, Joshi, Chen, Levy, Lewis, Zettlemoyer, and Stoyanov}]{liu}
Yinhan Liu, Myle Ott, Naman Goyal, Jingfei Du, Mandar Joshi, Danqi Chen, Omer Levy, Mike Lewis, Luke Zettlemoyer, and Veselin Stoyanov. 2020.
\newblock \href {https://arxiv.org/abs/1907.11692} {Roberta: A robustly optimized bert pretraining approach}.

\bibitem[{Maddison et~al.(2017)Maddison, Mnih, and Teh}]{maddison2017the}
Chris~J. Maddison, Andriy Mnih, and Yee~Whye Teh. 2017.
\newblock \href {https://openreview.net/forum?id=S1jE5L5gl} {The concrete distribution: A continuous relaxation of discrete random variables}.
\newblock In \emph{International Conference on Learning Representations}.

\bibitem[{Mao et~al.(2022)Mao, Mathias, Hou, Almahairi, Ma, Han, Yih, and Khabsa}]{mao-etal-2022-unipelt}
Yuning Mao, Lambert Mathias, Rui Hou, Amjad Almahairi, Hao Ma, Jiawei Han, Scott Yih, and Madian Khabsa. 2022.
\newblock \href {https://doi.org/10.18653/v1/2022.acl-long.433} {{U}ni{PELT}: A unified framework for parameter-efficient language model tuning}.
\newblock In \emph{Proceedings of the 60th Annual Meeting of the Association for Computational Linguistics (Volume 1: Long Papers)}, pages 6253--6264, Dublin, Ireland. Association for Computational Linguistics.

\bibitem[{Pfeiffer et~al.(2021)Pfeiffer, Kamath, R{\"u}ckl{\'e}, Cho, and Gurevych}]{pfeiffer-etal-2021-adapterfusion}
Jonas Pfeiffer, Aishwarya Kamath, Andreas R{\"u}ckl{\'e}, Kyunghyun Cho, and Iryna Gurevych. 2021.
\newblock \href {https://doi.org/10.18653/v1/2021.eacl-main.39} {{A}dapter{F}usion: Non-destructive task composition for transfer learning}.
\newblock In \emph{Proceedings of the 16th Conference of the European Chapter of the Association for Computational Linguistics: Main Volume}, pages 487--503, Online. Association for Computational Linguistics.

\bibitem[{Pfeiffer et~al.(2020)Pfeiffer, R{\"u}ckl{\'e}, Poth, Kamath, Vuli{\'c}, Ruder, Cho, and Gurevych}]{pfeiffer-etal-2020-adapterhub}
Jonas Pfeiffer, Andreas R{\"u}ckl{\'e}, Clifton Poth, Aishwarya Kamath, Ivan Vuli{\'c}, Sebastian Ruder, Kyunghyun Cho, and Iryna Gurevych. 2020.
\newblock \href {https://doi.org/10.18653/v1/2020.emnlp-demos.7} {{A}dapter{H}ub: A framework for adapting transformers}.
\newblock In \emph{Proceedings of the 2020 Conference on Empirical Methods in Natural Language Processing: System Demonstrations}, pages 46--54, Online. Association for Computational Linguistics.

\bibitem[{Poliak et~al.(2018)Poliak, Haldar, Rudinger, Hu, Pavlick, White, and Van~Durme}]{poliak-etal-2018-collecting}
Adam Poliak, Aparajita Haldar, Rachel Rudinger, J.~Edward Hu, Ellie Pavlick, Aaron~Steven White, and Benjamin Van~Durme. 2018.
\newblock \href {https://doi.org/10.18653/v1/D18-1007} {Collecting diverse natural language inference problems for sentence representation evaluation}.
\newblock In \emph{Proceedings of the 2018 Conference on Empirical Methods in Natural Language Processing}, pages 67--81, Brussels, Belgium. Association for Computational Linguistics.

\bibitem[{Radford et~al.(2018)Radford, Narasimhan, Salimans, and Sutskever}]{radford}
Alec Radford, Karthik Narasimhan, Tim Salimans, and Ilya Sutskever. 2018.
\newblock \href {https://www.mikecaptain.com/resources/pdf/GPT-1.pdf} {Improving language understanding by generative pre-training}.

\bibitem[{R{\"u}ckl{\'e} et~al.(2021)R{\"u}ckl{\'e}, Geigle, Glockner, Beck, Pfeiffer, Reimers, and Gurevych}]{ruckle-etal-2021-adapterdrop}
Andreas R{\"u}ckl{\'e}, Gregor Geigle, Max Glockner, Tilman Beck, Jonas Pfeiffer, Nils Reimers, and Iryna Gurevych. 2021.
\newblock \href {https://doi.org/10.18653/v1/2021.emnlp-main.626} {{AdapterDrop}: {O}n the efficiency of adapters in transformers}.
\newblock In \emph{Proceedings of the 2021 Conference on Empirical Methods in Natural Language Processing}, pages 7930--7946, Online and Punta Cana, Dominican Republic. Association for Computational Linguistics.

\bibitem[{Rudinger et~al.(2018)Rudinger, Naradowsky, Leonard, and Van~Durme}]{rudinger-etal-2018-gender}
Rachel Rudinger, Jason Naradowsky, Brian Leonard, and Benjamin Van~Durme. 2018.
\newblock \href {https://doi.org/10.18653/v1/N18-2002} {Gender bias in coreference resolution}.
\newblock In \emph{Proceedings of the 2018 Conference of the North {A}merican Chapter of the Association for Computational Linguistics: Human Language Technologies, Volume 2 (Short Papers)}, pages 8--14, New Orleans, Louisiana. Association for Computational Linguistics.

\bibitem[{Salemi et~al.(2023)Salemi, Mysore, Bendersky, and Zamani}]{salemi}
Alireza Salemi, Sheshera Mysore, Michael Bendersky, and Hamed Zamani. 2023.
\newblock \href {http://arxiv.org/abs/2304.11406} {La{MP}: When large language models meet personalization}.

\bibitem[{van~der Maaten and Hinton(2008)}]{maaten}
Laurens van~der Maaten and Geoffrey Hinton. 2008.
\newblock \href {https://www.jmlr.org/papers/volume9/vandermaaten08a/vandermaaten08a.pdf} {Visualizing data using t-sne}.
\newblock In \emph{Journal of Machine Learning Research 9 (2008)}.

\bibitem[{Wang et~al.(2019{\natexlab{a}})Wang, Pruksachatkun, Nangia, Singh, Michael, Hill, Levy, and Bowman}]{wang-superglue}
Alex Wang, Yada Pruksachatkun, Nikita Nangia, Amanpreet Singh, Julian Michael, Felix Hill, Omer Levy, and Samuel~R. Bowman. 2019{\natexlab{a}}.
\newblock \href {https://proceedings.neurips.cc/paper_files/paper/2019/hash/4496bf24afe7fab6f046bf4923da8de6-Abstract.html} {Super{GLUE}: A stickier benchmark for general-purpose language understanding systems}.
\newblock In \emph{Advances in Neural Information Processing Systems 32 (NeurIPS 2019)}.

\bibitem[{Wang et~al.(2019{\natexlab{b}})Wang, Singh, Michael, Hill, Levy, and Bowman}]{wang}
Alex Wang, Amanpreet Singh, Julian Michael, Felix Hill, Omer Levy, and Samuel~R. Bowman. 2019{\natexlab{b}}.
\newblock \href {https://arxiv.org/abs/1804.07461} {{GLUE}: A multi-task benchmark and analysis platform for natural language understanding}.
\newblock In \emph{International Conference on Learning Representations}.

\bibitem[{Wang et~al.(2022)Wang, Agarwal, Mukherjee, Liu, Gao, Awadallah, and Gao}]{wang-etal-2022-adamix}
Yaqing Wang, Sahaj Agarwal, Subhabrata Mukherjee, Xiaodong Liu, Jing Gao, Ahmed~Hassan Awadallah, and Jianfeng Gao. 2022.
\newblock \href {https://doi.org/10.18653/v1/2022.emnlp-main.388} {{A}da{M}ix: Mixture-of-adaptations for parameter-efficient model tuning}.
\newblock In \emph{Proceedings of the 2022 Conference on Empirical Methods in Natural Language Processing}, pages 5744--5760, Abu Dhabi, United Arab Emirates. Association for Computational Linguistics.

\bibitem[{Wolf et~al.(2020)Wolf, Debut, Sanh, Chaumond, Delangue, Moi, Cistac, Rault, Louf, Funtowicz, Davison, Shleifer, von Platen, Ma, Jernite, Plu, Xu, Le~Scao, Gugger, Drame, Lhoest, and Rush}]{wolf-etal-2020-transformers}
Thomas Wolf, Lysandre Debut, Victor Sanh, Julien Chaumond, Clement Delangue, Anthony Moi, Pierric Cistac, Tim Rault, Remi Louf, Morgan Funtowicz, Joe Davison, Sam Shleifer, Patrick von Platen, Clara Ma, Yacine Jernite, Julien Plu, Canwen Xu, Teven Le~Scao, Sylvain Gugger, Mariama Drame, Quentin Lhoest, and Alexander Rush. 2020.
\newblock \href {https://doi.org/10.18653/v1/2020.emnlp-demos.6} {Transformers: State-of-the-art natural language processing}.
\newblock In \emph{Proceedings of the 2020 Conference on Empirical Methods in Natural Language Processing: System Demonstrations}, pages 38--45, Online. Association for Computational Linguistics.

\bibitem[{Wu et~al.(2022)Wu, Chen, Xiao, Gu, and Sun}]{wu-etal-2022-pruning}
Jiarun Wu, Qingliang Chen, Zeguan Xiao, Yuliang Gu, and Mengsi Sun. 2022.
\newblock \href {https://doi.org/10.18653/v1/2022.findings-naacl.123} {Pruning adatperfusion with lottery ticket hypothesis}.
\newblock In \emph{Findings of the Association for Computational Linguistics: NAACL 2022}, pages 1632--1646, Seattle, United States. Association for Computational Linguistics.

\bibitem[{Zhou et~al.(2019)Zhou, Lan, Liu, and Yosinski}]{zhou}
Hattie Zhou, Janice Lan, Rosanne Liu, and Jason Yosinski. 2019.
\newblock \href {https://proceedings.neurips.cc/paper_files/paper/2019/hash/1113d7a76ffceca1bb350bfe145467c6-Abstract.html} {Deconstructing lottery tickets: Zeros, signs, and the supermask}.
\newblock In \emph{Advances in Neural Information Processing Systems 32 (NeurIPS 2019)}.

\end{thebibliography}

\newpage
\appendix
\section{Algorithms} \label{sec:app_alg}

We suggest the hard softmax algorithm employing straight-through gradient estimation, as outlined in Algorithm \ref{alg:softmax}.

\begin{algorithm}
\caption{Hard (top-$k$) Softmax (Straight-Through Gradient Estimation)}\label{alg:softmax}
\begin{algorithmic}
\Require Input \texttt{logits}, noise level \texttt{nu}, temperature \texttt{tau} and \texttt{k} for top-$k$ selection
\Ensure Vector \texttt{y} representing the top-$k$ elements
\State \texttt{logits = logits + nu * Gumbel(0, 1)}
\State \texttt{y\_soft = softmax(logits / tau)}
\State \texttt{indices = topk(y\_soft)}
\State \texttt{y\_hard = khot\_encoding(indices)}
\State \texttt{y\_hard = y\_hard / k}
\State \texttt{y = y\_hard - y\_soft.detach() + y\_soft}
\end{algorithmic}
\end{algorithm}

\section{Figures} \label{sec:app_fig}

\begin{figure}[h]
    \centering
    \ifnum\usesvg=0
        \includegraphics[width=1.0\linewidth]{fig_random_seed.png}
    \else
        \def\svgwidth{\columnwidth}
        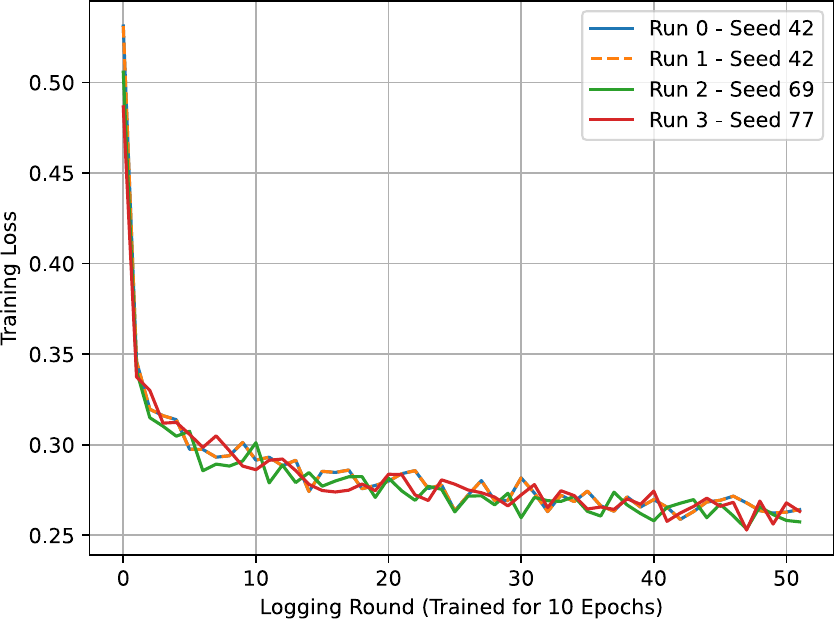
    \fi
    \caption{\textbf{Training loss curves for \texttt{sst2} ($N = 100$, soft) with varying random seeds.} While local fluctuations in the loss curves differ from each other, they tend to follow a similar trajectory globally. It's evident that our experiments guarantee reproducibility, as two different runs with random seed 42 yield identical loss curves. (The blue solid line and the orange dashed line are completely overlapped.) In terms of evaluation scores, run 0 and 1 both recorded 0.8956, run 3 recorded 0.8865, and run 4 recorded 0.8968, with little variation among them.}
    \label{fig:random_seed}
\end{figure}

Figure \ref{fig:random_seed} illustrates our experiments with different random seeds, revealing consistent trends in the results. It also includes two runs with the same random seed, demonstrating the reproducibility of our experiments based on the random seed.

\section{Hyper-Parameters}

The major hyper-parameters are as follows:
\begin{itemize}
    \setlength\itemsep{0.0em}
    \item $N$ (The number of adapters attached to an X-PEFT model): Generally, the more adapters, the better the performance. However, considering training budgets, 100, 150, or 200 adapters are quite good choices. Our search space for $N$ was $\{100, 200, 400, 800\}$.
    \item $k$ ($k$ for top-$k$ selection for hard masking): We found that $k = 50$ is a reasonably good choice regardless of other settings, as discussed in the ablation studies. Our search space for $k$ was $\{1, 10, 20, 30, 40, 50, 60, 70, 80, 100\}$.
    \item $b$ (Bottleneck dimension of adapters used in an X-PEFT model): The bottleneck dimension of adapters used in an X-PEFT model has no significant impact on the model's performance. We used a default value (48) provided by AdapterHub. Our search space for $b$ was $\{12, 24, 48, 96\}$.
    \item Batch size: The batch size has no significant impact on the model's performance. A batch size of 64 is suitable for our technical environment. Our search space for batch size was $\{8, 16, 32, 64, 128\}$.
\end{itemize}

We used \texttt{bert-base-uncased}, so the following hyperparameters were consistent across all experiments:

\begin{itemize}
    \setlength\itemsep{0.0em}
    \item $L$ (The number of blocks of the PLM)
    \item $d$ (Input dimension into adapter layers)
\end{itemize}

\section{Modification Details for the LaMP dataset} \label{sec:app_lamp_mod}

In our research, we utilized the LaMP-2 dataset, specifically the `Personalized News Categorization' dataset, which is part of the LaMP benchmark \citep{salemi}. However, several modifications were necessary to adapt this dataset for our specific purposes.

The original LaMP-2 dataset was primarily designed for prompt tuning, aiming to understand how specific authors categorize given news articles. Each data point in this dataset consisted of the news article text and the author's profile. It's essential to clarify that the author's profile is not an identifier but rather a collection of news article texts authored by that particular individual, along with the categories assigned by the author to these articles.

As our experiments focused on standard supervised classification, we needed datasets containing pairs of news texts and their corresponding labels, alongside the author's identity. In other words, our data schema needed to be in the format of (\texttt{news\_text}, \texttt{news\_category}, \texttt{author\_id}). Here, \texttt{author\_id} simply refers to a numerical identifier that can also be used as a label.

To meet these requirements, we exclusively extracted the author profile data from the original LaMP-2 dataset and proceeded to modify it according to the specified format. Given that the same author's data may appear more than once in the original LaMP-2 dataset, we took care to remove any duplicates in our modified version.

Out of the 8,090 data points in the LaMP-2 dataset, we extracted 17,005 news texts, each categorized into one of 15 categories, authored by 323 unique authors, eliminating any duplicates.

\section{Training Time}

Information regarding the training time for our GLUE and SuperGLUE experiments can be found in Table \ref{tab:comp_cost1} and Table \ref{tab:comp_cost2}.

Here are the training times for our LaMP experiments:
\begin{itemize}
    \setlength\itemsep{0.0em}
    \item \texttt{x\_peft warm} (hard): 5.06 hours
    \item \texttt{x\_peft warm} (soft): 4.90 hours
    \item \texttt{x\_peft random} (hard): 8.36 hours
    \item \texttt{x\_peft random} (soft): 8.40 hours
\end{itemize}

\section{Trained Parameters}

The parameter count of \texttt{bert-base-uncased} is known to be 110M. All the X-PEFT configurations that we used in the experiments and their parameter counts including \texttt{bert-base-uncased} is as follows (with $c$ representing the label count for a downstream head):

\begin{itemize}
    \setlength\itemsep{0.0em}
    \item $N = 100$ and $c = {2, 3, 15}$: 200M
    \item $N = 150$ and $c = {2, 3, 15}$: 245M
    \item $N = 200$ and $c = {2, 3, 15}$: 290M
    \item $N = 400$ and $c = {2, 3, 15}$: 468M
    \item $N = 800$ and $c = {2, 3, 15}$: 826M
\end{itemize}

The counts of trained parameters, both including and excluding the downstream head, are provided in Table \ref{tab:app_param_count_head}.

\begin{table*}[]
    \centering
    \begin{tabular}{ccccc}
    \hline
    & \multicolumn{3}{c}{\textbf{Including Head}} & \textbf{Excluding Head}\\
    \textbf{$N$} & \textbf{$c=2$} & \textbf{$c=3$} & \textbf{$c=15$} &\\
    \hline
    100 & 0.596M & 0.596M & 0.606M & 0.004M \\
    150 & 0.597M & 0.598M & 0.607M & 0.005M \\
    200 & 0.598M & 0.599M & 0.608M & 0.006M \\
    400 & 0.603M & 0.604M & 0.613M & 0.011M \\
    800 & 0.612M & 0.613M & 0.622M & 0.020M \\
    \hline
    \end{tabular}
    \caption{\textbf{Trained parameter count including and excluding head.} $c$ denotes the label count for a downstream head.}
    \label{tab:app_param_count_head}
\end{table*}

\section{Detailed GLUE and SuperGLUE Evaluations} \label{sec:app_eval}

Here are the complete evaluations for the GLUE and SuperGLUE benchmarks. Refer to Table \ref{tab:glue_eval_supp1}, Table \ref{tab:glue_eval_supp2} and Table \ref{tab:superglue_eval_supp}.

\begin{table*}
    \centering
    \begin{tabular}{cccccccc}
        \hline
        \textbf{Mode} & \textbf{Adapters} & \textbf{\texttt{cola}} & \textbf{\texttt{sst2}} & \textbf{\texttt{mrpc}} & \textbf{\texttt{mrpc}} & \textbf{\texttt{qqp}} & \textbf{\texttt{qqp}}\\
        & & \scriptsize{(MCC)} & \scriptsize{(Acc)} & \scriptsize{(Acc)} & \scriptsize{(F1)} & \scriptsize{(Acc)} & \scriptsize{(F1)}\\
        \hline
        \texttt{x\_peft} & 100 \small{(soft)} & 0.3977 & 0.8956 & 0.7353 & 0.8291 & 0.8132 & 0.7643\\
        & 100 \small{(hard)} & 0.3891 & 0.8716 & 0.7132 & 0.8146 & 0.7824 & 0.7307\\
        & 200 \small{(soft)} & 0.4422 & 0.9106 & 0.7328 & 0.8278 & 0.8266 & 0.7793\\
        & 200 \small{(hard)} & 0.4446 & 0.8911 & 0.7745 & 0.8521 & 0.7933 & 0.7480\\
        & 400 \small{(soft)} & 0.4654 & 0.8991 & 0.7328 & 0.8250 & 0.8345 & 0.7845\\
        & 400 \small{(hard)} & 0.4592 & 0.8899 & 0.7843 & 0.8562 & 0.8011 & 0.7515\\
        \hdashline
        \texttt{head\_only} & - & 0.3122 & 0.8521 & 0.7059 & 0.8187 & 0.7575 & 0.6884\\
        \hdashline
        \texttt{single\_adapter} & - & 0.4277 & 0.9140 & 0.7034 & 0.8130 & 0.8688 & 0.8263\\
        \hline
    \end{tabular}    
    \caption{\textbf{Evaluation of the GLUE tasks (part 1).} In the case of hard masking, we employ $k = 50$ for top-$k$ selection. The scores in the table are reported based on the official metrics provided by the GLUE dataset. `Acc,' `MCC,' and `F1' denote accuracy, Matthew's Correlation, and F1 score, respectively.
    }
    \label{tab:glue_eval_supp1}    
\end{table*}

\begin{table*}
    \centering
    \begin{tabular}{ccccccccc}
        \hline
        \textbf{Mode} & \textbf{Adapters} & \textbf{\texttt{stsb}} & \textbf{\texttt{stsb}} & \textbf{\texttt{mnli}} & \textbf{\texttt{mnli}} & \textbf{\texttt{qnli}} & \textbf{\texttt{rte}} & \textbf{\texttt{wnli}}\\
        & & \scriptsize{(PCC)} & \scriptsize{(SRC)} & \scriptsize{(Acc)} & \scriptsize{(AMM)} & \scriptsize{(Acc)} & \scriptsize{(Acc)} & \scriptsize{(Acc)}\\
        \hline
        \texttt{x\_peft} & 100 \small{(soft)} & 0.7888 & 0.7948 & 0.6663 & 0.6894 & 0.8182 & 0.5776 & 0.3380\\
        & 100 \small{(hard)} & 0.7404 & 0.7492 & 0.6186 & 0.6372 & 0.7626 & 0.6101 & 0.3239\\
        & 200 \small{(soft)} & 0.8001 & 0.8076 & 0.6863 & 0.7013 & 0.8343 & 0.5957 & 0.3662\\
        & 200 \small{(hard)} & 0.7506 & 0.7646 & 0.6320 & 0.6597 & 0.7891 & 0.5776 & 0.3380\\
        & 400 \small{(soft)} & 0.8028 & 0.8089 & 0.7074 & 0.7275 & 0.8349 & 0.5848 & 0.2958\\
        & 400 \small{(hard)} & 0.8115 & 0.8148 & 0.6569 & 0.6789 & 0.8083 & 0.5487 & 0.2676\\
        \hdashline
        \texttt{head\_only} & - & 0.4687 & 0.4482 & 0.5307 & 0.5335 & 0.6842 & 0.5884 & 0.3803\\
        \hdashline
        \texttt{single\_adapter} & - & 0.7995 & 0.8057 & 0.7934 & 0.8034 & 0.8812 & 0.5993 & 0.4225\\
        \hline
    \end{tabular}
    \caption{\textbf{Evaluation of the GLUE tasks (part 2).} In the case of hard masking, we employ $k = 50$ for top-$k$ selection. The scores in the table are reported based on the official metrics provided by the GLUE dataset. `Acc,' `PCC,' and `SRC' denote accuracy, Pearson correlation, and Spearman correlation, respectively. For \texttt{mnli}, `Acc' and `AMM' denote accuracy matched and accuracy mismatched, respectively.
    }
    \label{tab:glue_eval_supp2}    
\end{table*}

\begin{table*}
    \centering
    \begin{tabular}{ccccccc}
        \hline
        \textbf{Mode} & \textbf{Adapters} & \textbf{\texttt{cb}} & \scalebox{0.8}{\textbf{\texttt{boolq}}} & \textbf{\texttt{axb}} & \textbf{\texttt{axg}} & \textbf{\texttt{axg}}\\
        & & \scriptsize{(Acc)} & \scriptsize{(Acc)} & \scriptsize{(MCC)} & \scriptsize{(Acc)} & \scriptsize{(GPS)}\\
        \hline
        \texttt{x\_peft} & 100 \small{(soft)} & 0.6429 & 0.6676 & 0.1111 & 0.5253 & 92.6724\\
        & 100 \small{(hard)} & 0.6786 & 0.6569 & 0.0943 & 0.4831 & 86.6379\\
        & 200 \small{(soft)} & 0.6786 & 0.6599 & 0.0721 & 0.5197 & 96.1207\\
        & 200 \small{(hard)} & 0.6786 & 0.6648 & 0.0244 & 0.5028 & 88.3621\\
        & 400 \small{(soft)} & 0.6786 & 0.6599 & 0.0916 & 0.5084 & 93.5345\\
        & 400 \small{(hard)} & 0.6964 & 0.6792 & 0.1203 & 0.5000 & 94.8276\\
        \hdashline
        \texttt{head\_only} & - & 0.7143 & 0.6358 & 0.0869 & 0.4972 & 82.3276\\
        \hdashline
        \texttt{single\_adapter} & - & 0.6786 & 0.6489 & 0.1027 & 0.5084 & 93.5345\\
        \hline
    \end{tabular}
    \caption{\textbf{Evaluation of the SuperGLUE tasks.} In the case of hard masking, we employ $k = 50$ for top-$k$ selection. `GPS' denotes Gender Parity Score, and all other symbols can be understood in the same context as in Table \ref{tab:glue_eval_supp1} and \ref{tab:glue_eval_supp2}
    }
    \label{tab:superglue_eval_supp}
\end{table*}

\begin{table*}[ht!]
    \centering
    \begin{tabular}{ccccccccccccc}
        \hline
        \textbf{Mode} & \textbf{Adapters} & \textbf{\texttt{cola}} & \textbf{\texttt{sst2}} & \textbf{\texttt{mrpc}} & \textbf{\texttt{qqp}} & \textbf{\texttt{stsb}} & \textbf{\texttt{mnli}} & \textbf{\texttt{qnli}}  & \textbf{\texttt{rte}} & \textbf{\texttt{wnli}}\\
        \hline
        \texttt{x\_peft} & 100 \small{(soft)} & 0.55 & 4.32 & 0.25 & 26.07 & 0.38 & 24.20 & 6.71 & 0.17 & 0.05\\
        & 100 \small{(hard)} & 0.57 & 6.12 & 0.26 & 26.69 & 0.38 & 24.32 & 7.02 & 0.17 & 0.05\\
        & 200 \small{(soft)} & 1.11 & 8.10 & 0.48 & 43.67 & 0.71 & 47.12 & 12.61 & 0.32 & 0.10\\
        & 200 \small{(hard)} & 1.11 & 8.51 & 0.50 & 44.13 & 0.71 & 47.33 & 12.54 & 0.33 & 0.09\\
        & 400 \small{(soft)} & 2.16 & 16.93 & 0.92 & 90.43 & 1.40 & 104.57 & 29.29 & 0.62 & 0.19\\
        & 400 \small{(hard)} & 2.07 & 16.91 & 0.91 & 91.45 & 1.41 & 108.14 & 26.15 & 0.78 & 0.19\\
        \hdashline
        \texttt{head\_only} & - & 0.04 & 0.47 & 0.02 & 1.08 & 0.04 & 2.70 & 0.46 & 0.01 & 0.00\\
        \hdashline
        \texttt{single\_adapter} & - & 0.09 & 0.55 & 0.03 & 2.97 & 0.05 & 4.22 & 1.61 & 0.01 & 0.01\\    
        \hline
    \end{tabular}
    \caption{{\bf Computation cost of the GLUE tasks (training time, hours).}
    }
    \label{tab:comp_cost1}
\end{table*}

\begin{table*}[t]
    \centering
    \begin{tabular}{ccccccc}
        \hline
        \textbf{Mode} & \textbf{Adapters} & \textbf{\texttt{cb}} & \scalebox{0.8}{\textbf{\texttt{boolq}}} & \textbf{\texttt{axb}} & \textbf{\texttt{axg}}\\
        \hline        
        \texttt{x\_peft} & 100 \small{(soft)} & 0.02 & 0.60 & 0.18 & 0.18\\
        & 100 \small{(hard)} & 0.02 & 0.61 & 0.18 & 0.18\\
        & 200 \small{(soft)} & 0.03 & 1.17 & 0.33 & 0.33\\
        & 200 \small{(hard)} & 0.03 & 1.19 & 0.35 & 0.35\\
        & 400 \small{(soft)} & 0.06 & 2.29 & 0.64 & 0.64\\
        & 400 \small{(hard)} & 0.06 & 2.41 & 0.68 & 0.68\\
        \hdashline
        \texttt{head\_only} & - & 0.00 & 0.07 & 0.02 & 0.02\\
        \hdashline
        \texttt{single\_adapter} & - & 0.00 & 0.08 & 0.03 & 0.03\\
        \hline
    \end{tabular}
    \caption{{\bf Computation cost of the SuperGLUE tasks (training time, hours).}}
    \label{tab:comp_cost2}
\end{table*}

\end{document}